%% file: main_arxiv.tex
\documentclass[reprint, aps, pre, superscriptaddress, floatfix]{revtex4-2}
\usepackage{bm}
\usepackage{overpic}
\usepackage[normalem]{ulem}
\usepackage{xcolor}
\usepackage{comment}
\definecolor{blue(ncs)}{rgb}{0.0, 0.53, 0.74}

\definecolor{truc(ncs)}{rgb}{0.5, 0.53, 0.74}
\definecolor{violet(ncs)}{rgb}{0.58, 0.0, 0.83}

\definecolor{red(ncs)}{rgb}{0.90, 0.40, 0.40}

\newcommand{\figpanel}[1]{\footnotesize(\textbf{{#1}})}
\input{math_commands}

\begin{document}

\title{Equilibrium Training of Energy-Based Models with Parallel Trajectory Tempering}

\author{Nicolas Béreux}
\affiliation{Université Paris-Saclay, CNRS, INRIA,
LISN, 91190 Gif-sur-Yvette, France}
\author{Aurélien Decelle}
\affiliation{Escuela Técnica Superior de Ingenieros Industriales, Universidad Politécnica de Madrid, 28006 Madrid, Spain}
\author{Cyril Furtlehner}
\affiliation{Université Paris-Saclay, CNRS, INRIA,
LISN, 91190 Gif-sur-Yvette, France}
\author{Beatriz Seoane}
\affiliation{Departamento de Física Teórica  \& IPARCOS, Universidad Complutense de Madrid, 28040 Madrid, Spain.}

\begin{abstract}
Energy-Based Models (EBMs) provide an interpretable framework for generative modeling of scientific data, but poor Markov Chain Monte Carlo mixing often limits their reliability. We introduce a training algorithm based on Parallel Trajectory Tempering (PTT), which exploits the continuity of the optimization path to maintain equilibrium sampling throughout learning. This enables stable and fast training on highly multimodal and data-scarce scientific datasets.
Combined with reservoir sampling and adaptive optimization, PTT has a computational cost comparable to Persistent Contrastive Divergence, making it a practical replacement for standard training methods. It also provides direct estimates of thermalization times, equilibrium samples from trained models, and accurate log-likelihoods at essentially no additional cost.
Experiments on Restricted Boltzmann Machines show that PTT consistently outperforms existing EBM training approaches. On discrete tabular data, it also surpasses state-of-the-art deep generative models, yielding higher-quality samples and greater robustness to overfitting and limited data. Our results make equilibrium maximum-likelihood training of EBMs practical and computationally efficient.
\end{abstract}

\maketitle

Advances in imaging, genome sequencing, and high-throughput experiments are generating vast high-dimensional datasets across biological scales, from molecules and cells to neural circuits and populations. These data create new opportunities to study living and cognitive systems as many-body systems shaped by collective behavior, disorder, fluctuations, and non-equilibrium dynamics. Realizing this potential requires generative models that combine sufficient expressivity to capture complex dependencies with enough interpretability to reveal the organizing principles of biological systems.

Modern generative models have achieved remarkable success in modeling high-dimensional data, but many do not provide explicit access to the probability distributions they learn, limiting their value for scientific discovery, while remaining largely hard to interpret. In contrast, Energy-Based Models (EBMs) represent the data distribution as a Boltzmann distribution,
\begin{equation}\label{eq:pdf}
p_{\bm\theta}(\bm v)=\frac{\exp\left[-E_{\bm\theta}(\bm v)\right]}{Z_{\bm\theta}},
\qquad
Z_{\bm\theta}=\sum_{\{\bm v\}}\exp\left[-E_{\bm\theta}(\bm v)\right],
\end{equation}
where $E_{\bm\theta}(\bm v)$ is a neural-network energy and $(Z_{\bm\theta})$ the partition function. EBMs connect directly to Boltzmann statistical mechanics~\cite{hinton2025nobel}: learning infers an effective many-body Hamiltonian whose low-free-energy states encode dominant data structures. Since any energy function can be exactly decomposed into irreducible interactions of increasing order~\cite{bulso2021restricted}, EBMs provide a principled route to higher-order dependencies~\cite{decelle2025inferring,decelle2026distributional}. For shallow EBMs, the energy landscape can also be tracked during training~\cite{tubiana2017emergence,tubiana2019learning,decelle2023unsupervised}, enabling principled pattern extraction and interpretable analyses of experimental data~\cite{di2025unbearable}, with applications in computational biology~\cite{tubiana2017emergence,bravi2021rbm,bravi2024development}, neuroscience~\cite{van2023neural,bereux2026uncovering,dommanget2026cross}, statistical physics~\cite{melko2019restricted,barra2018phase,decelle2021restricted}, and quantum physics~\cite{carleo2017solving,nomura2017restricted}. Despite these advantages, EBMs have been largely overshadowed in modern machine learning, not for lack of expressive power, but because they remain difficult to train~\cite{decelle2021equilibrium,liao2022gaussian}.

Maximum-likelihood learning requires model samples, typically generated by MCMC. As training progresses, EBMs develop increasingly structured energy landscapes and undergo cascades of second-order phase transitions~\cite{Decelle_2017,decelle2018thermodynamics,bachtis2024cascade}, whose critical slowing down makes equilibrium sampling and reliable model evaluation increasingly difficult~\cite{krause_algorithms_2020}. Training therefore often relies on out-of-equilibrium chains, yielding biased gradients, unstable optimization, and poor generalization, particularly for high-dimensional and multimodal data~\cite{nijkamp2019learning,nijkamp2020anatomy,decelle2021equilibrium,agoritsas2023explaining,bereux2023learning,carbone2024fast,bereux2025fast}. Contrastive Divergence (CD)~\citep{hinton2002training} made EBM training practical by replacing equilibrium sampling with short chains, while Persistent Contrastive Divergence (PCD)~\citep{tieleman2008training} improves on this by maintaining chains across updates.

Yet, on structured datasets, persistent chains may eventually fail to track the evolving model distribution and fall out of equilibrium, thereby biasing the likelihood gradient~\citep{bereux2023learning,bereux2025fast}. Similar failures arise in data-scarce regimes, where persistent chains become trapped near individual training examples, causing mixing times to diverge and training to break down abruptly, as illustrated for MNIST in the Appendix. Such failure modes are often overlooked in likelihood-based comparisons, despite their strong impact on both model behavior and generative quality: poorly trained models typically exhibit anomalously slow relaxation, increased overfitting, memorization and mode collapse effects, and unreliable evaluation. 
Controlling equilibration throughout training is therefore particularly important in scientific applications, where interpretability is a central advantage of EBMs.

Low-rank pretraining can delay the breakdown of persistent chains~\citep{decelle2021exact,bereux2025fast}, while constrained MCMC~\citep{bereux2023learning}, population annealing~\citep{krause2018population}, reweighting~\citep{carbone2024efficient} improve exploration at the cost of substantial computation or reduced effective sample size. Parallel Tempering (PT)~\citep{marinari1992simulated,lyubartsev1992new,geyer1995annealing,tesi1996monte,hukushima1996exchange}  where the system is replicated at a given lader of temperatures $\{\beta_t\}$, exchanges replicas across neighboring temperatures and is effective on many datasets~\citep{salakhutdinov2009learning,desjardins2010tempered,krause_algorithms_2020}, but requires many intermediate temperatures and fails in highly structured landscapes, where first-order transitions suppress exchanges and trap replicas~\citep{decelle2021exact,bereux2025fast}.

\begin{figure}
    \centering
    \includegraphics[width=0.5\linewidth]{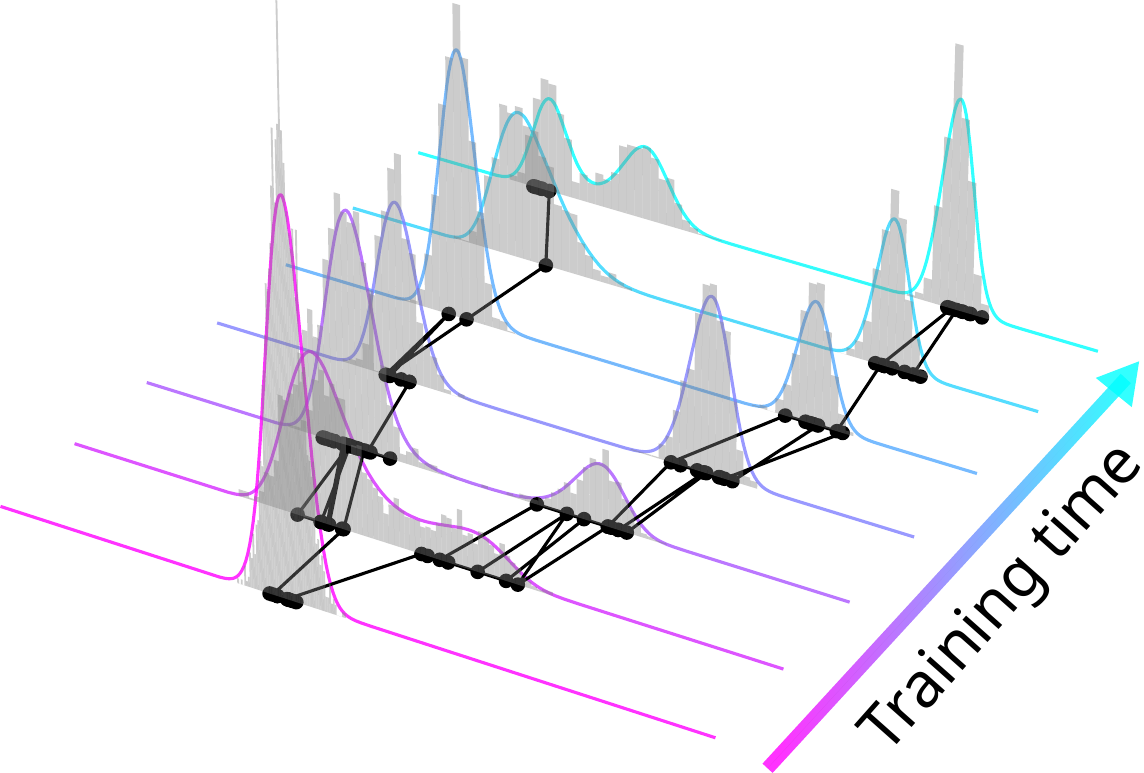}
    \caption{Schematic illustration of the Parallel Trajectory Tempering (PTT) algorithm. As training progresses (arrow), successive model checkpoints are added to the ladder. Replicas perform a random walk between neighboring checkpoints via swap moves, ensuring efficient equilibration despite the increasingly structured energy landscape.}
    \label{fig:plot_ptt}
\end{figure}
Motivated by a theoretical analysis of RBM learning dynamics, Parallel Trajectory Tempering (PTT) was recently introduced~\citep{bereux2025fast}. Rather than tempering in temperature, PTT exchanges replicas between neighboring models along the learning trajectory (Fig.~\ref{fig:plot_ptt}), following Hamiltonian-exchange MCMC~\citep{rosta2011catalytic}. In this work, we make PTT practical for EBM training by combining reservoir sampling with adaptive optimization, achieving a computational cost comparable to PCD while maintaining equilibrium sampling even for highly structured or very small datasets. This largely removes the main bottleneck in EBM training—the negative phase—and enables equilibrium training, sampling, and evaluation at near-standard computational cost.

\section{Background}
\subsection{Maximum-likelihood training}

EBMs are typically trained by maximizing the log-likelihood (LL) of a training dataset
$\mathcal{D}=\{\bm v^{(\mu)}\}_{\mu=1}^M$, where $M$ is the number of training examples.
The objective function is
\begin{equation}
\mathcal{L}=\mean{\log p_{\bm\theta}(\bm v)}_{\mathcal D},
\end{equation}
where $\mean{\cdot}_{\mathcal D}$ denotes the empirical average over the empirical training distribution
$
p_{\mathcal D}(\bm v)=\frac{1}{M}\sum_{\mu=1}^M\delta\!\left(\bm v-\bm v^{(\mu)}\right).
$
Starting from an initial set of parameters, typically chosen at random, the model parameters are updated by gradient ascent on the log-likelihood:
\begin{equation}
  \nabla_{\bm\theta} \mathcal L
  =
  \mean{-\nabla_{\bm\theta}E}_{\mc D}
  -
  \mean{-\nabla_{\bm\theta}E}_{\mc M},
\end{equation}
where $\mean{\cdot}_{\mc M}$ denotes the average with respect to the model distribution [\ref{eq:pdf}]. The parameters are then updated according to
\begin{equation}
\bm\theta \leftarrow \bm\theta + \eta \nabla_{\bm\theta}\mathcal L,
\end{equation}
where $\eta$ is the learning rate.
The data term, $\mean{-\nabla_{\bm\theta}E}_{\mc D}$, is directly estimated from the training set, whereas the model term requires averaging over an exponentially large state space and is therefore approximated by MCMC. This estimate is unbiased only after equilibration, but the thermalization time of Alternating Gibbs Sampling (AGS) typically grows rapidly during training~\cite{decelle2021equilibrium,bereux2023learning}, making equilibrium maximum-likelihood learning impractical and motivating heuristic approaches such as CD and PCD.

\subsection{Restricted Boltzmann Machines (RBMs)} In this work, we use RBMs~\cite{Smolensky,hinton2025nobel} as a paradigmatic EBM. Despite their simple bipartite architecture, RBMs are universal approximators~\cite{le2008representational} and capture high-order dependencies between visible variables through binary hidden units~\cite{decelle2024inferring}. The absence of intra-layer couplings makes AGS particularly efficient. The RBM joint energy reads
\begin{equation}
\mathcal{H}_{\bm\theta}(\boldsymbol{v},\boldsymbol{h})
=
-\sum_{i,\mu}W_{i\mu}v_i h_\mu
-\sum_i a_i v_i
-\sum_\mu b_\mu h_\mu .
\end{equation}
With $\bm \theta=\{W,\bm a,\bm b\}$
After marginalizing the binary hidden variables, one obtains the RBM energy function
$p_{\bm\theta}(\boldsymbol{v}) \!=\!\frac{1}{Z}\sum_{\bm h}e^{-\mathcal{H}_{\bm\theta}(\bm v,\bm h)}\!=\!\frac{1}{Z}e^{-E_{\bm\theta}(\boldsymbol{v})}$, with
\begin{equation}
\!\!E_{\bm \theta}(\boldsymbol{v})
\!=
\!-\!\sum_i a_i v_i
-\!\sum_{\mu=1}^{N_\mathrm{h}}
\log\!\left[\!1\!+\!\exp\left(\!b_\mu\!+\!\!\sum_i W_{i\mu}v_i\!\right)\!\right]\!\!.
\end{equation}
Despite their simplicity, RBMs retain the main sampling challenges of EBMs, including multimodality, slow mixing, and training-induced phase transitions. Yet, as shown below, they achieve strong generative performance on discrete and mixed-type tabular data, outperforming state-of-the-art methods scientific datasets.

\section{New training method}

Recent work~\cite{bereux2023learning} introduced Parallel Trajectory Tempering (PTT), exploiting the smooth evolution of the model distribution during training. Rather than tempering in temperature, PTT uses a ladder of model checkpoints $\{E_t\}$ (also called replica) along the training trajectory, interpolating between initialization and the trained model (see sketch in Fig.~\ref{fig:plot_ptt}. Configurations are exchanged between neighboring replicas according to the Metropolis rule
\begin{equation}
 \! \textstyle p_{\mathrm{acc}}(\bm{x}_t \!\leftrightarrow\! \bm{x}_{t-1})\! = \!\min \left[1, e^{\Delta E_t(\bm{x}_t)\! - \!\Delta E_t(\bm{x}_{t-1})}  \right]\!\! \label{eq:PTexch}
\end{equation}
where $\Delta E_t(\bm{x}) \equiv E_t(\bm{x}) - E_{t-1}(\bm{x})$. In practice, each PTT sweep combines one swap proposal between nearby replicas followed by $k$ AGS steps in each model.

Because PTT exploits the smooth evolution of the model distribution during training, the replica ladder can be built adaptively. Starting from the random initialization, a checkpoint is frozen whenever the swap acceptance with the previous replica falls below a threshold $\alpha$, ensuring sufficient overlap between neighbors. If this occurs after a single gradient update, the learning rate is temporarily reduced; a cosine-similarity criterion is used to prevent it from becoming excessively small (see Appendix).

A naive implementation would simulate all replicas, with a cost growing linearly with ladder size. We avoid this by introducing a \emph{reservoir} of $N_{\rm res}$ equilibrium samples. When a new checkpoint $t$ is added, replica $t-1$ is thermalized for more than $20\tau_{\mathrm{exp}}$, after which samples are stored every $2\tau_{\mathrm{int}}$ until the reservoir is filled. Thereafter, only replicas from $t$ onward are evolved, while exchanges with $t-1$ use configurations drawn from the reservoir. Since new checkpoints are created only $O(10)$ times during training, the full ladder is rarely simulated; most updates require evolving only the last two replicas.

Beyond substantially accelerating sampling, PTT provides, essentially for free, an accurate estimate of the log-likelihood throughout and after training~\cite{bereux2025fast}, once the replica ensemble has equilibrated. This is achieved by recursively evaluating the partition function,
\begin{equation}\label{eq:ZPTT}
Z_t
=
\Big\langle e^{E_{t-1}-E_t}\Big\rangle_{t-1}\,Z_{t-1},
\end{equation}
where $\langle\cdot\rangle_t$ denotes an expectation over the equilibrium distribution of replica $t$. Since these expectations are computed from the same Markov chains used for training, or to sample the different models after training using PTT, no additional computation cost is required. PTT also provides a direct thermalization diagnostic from replica diffusion across the ladder~\cite{alvarez2010nature}. At equilibrium, each replica visits the $N_\mathrm{m}$ models uniformly, while the checkpoint-index of each replica autocorrelation function, $C(t)\sim e^{-t/\tau_{\mathrm{exp}}}$, yields the relaxation time $\tau_{\mathrm{exp}}$ and, through its integral, $\tau_{\mathrm{int}}$, setting the $2\tau_{\mathrm{int}}$ spacing between statistically independent samples (see Appendix).

\section{Results}
We compare PTT with standard PCD for RBM training and with state-of-the-art generative models, including Bayesian Flow Networks (BFNs)~\cite{graves2023bayesian}, across diverse scientific datasets: (i) the 2D Ising model, (ii) mouse neural spike recordings, (iii) homologous protein families, (iv) medical patient data, and (v) low-data image datasets. These problems combine strong multimodality, limited sample sizes, and strong temporal correlations. In such regimes, PCD chains can remain trapped in individual modes, preventing accurate estimation of their relative weights. We show that PTT-trained RBMs achieve higher test log-likelihoods, generate more diverse and higher-quality samples, are easier to evaluate and sample from than PCD-trained models, and display markedly greater robustness to overfitting. They also outperform competing generative models across these scientific datasets.

For all results, $40\%$ of the data was held out as a test set to assess generalization and detect overfitting, and models were selected at the checkpoint with the highest test log-likelihood. When generating samples from the trained RBMs, thermalization was systematically verified before saving configurations, following the protocol described in Appendix.

\begin{figure}
    \centering
    \begin{overpic}[width=\linewidth]{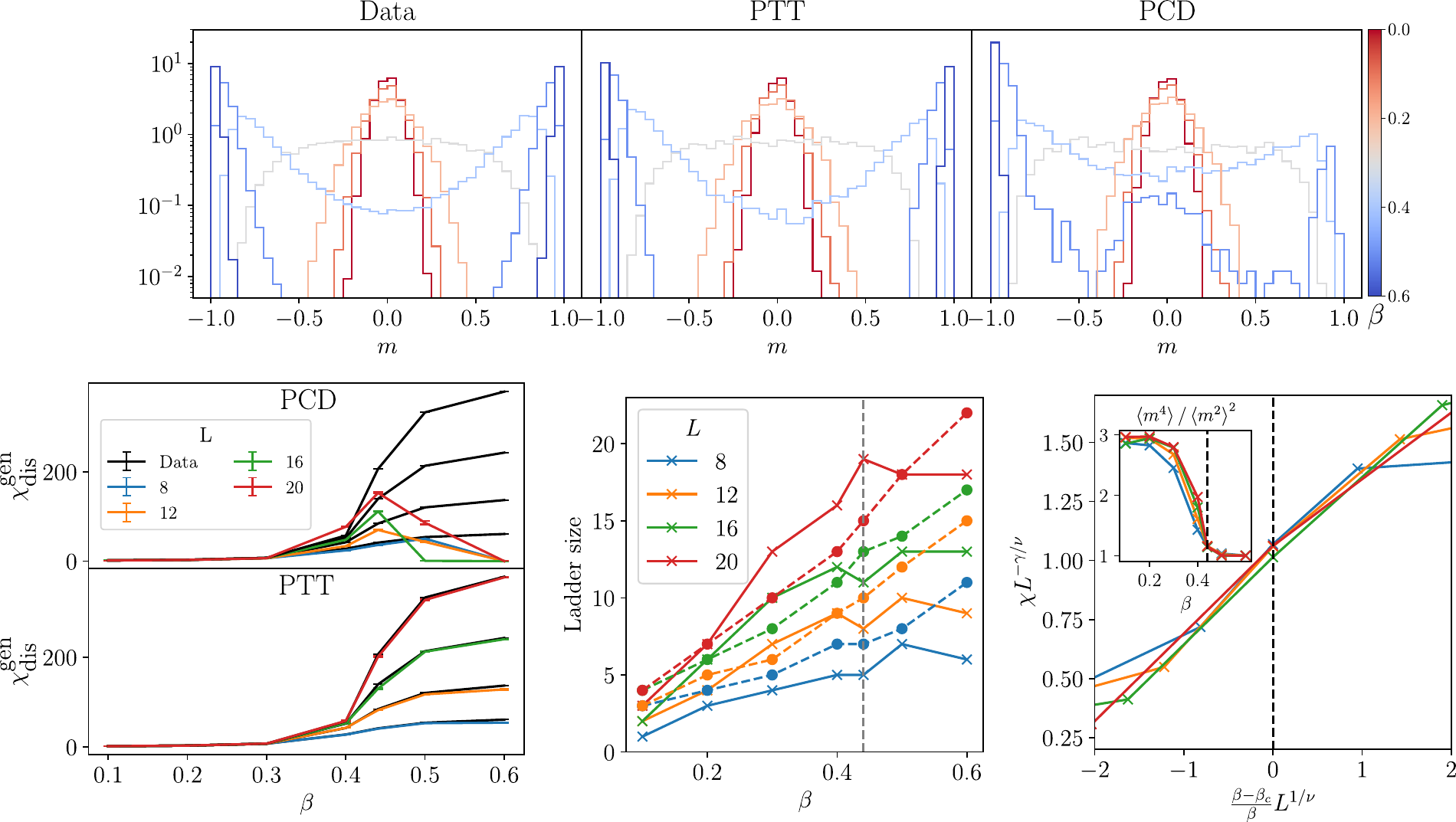}
    \put(4,55){{\figpanel{a}}}
    \put(0,30){{\figpanel{b}}}
    \put(37,30){{\figpanel{c}}}
    \put(69,30){{\figpanel{d}}}
    \end{overpic}
\caption{\textbf{Learning the 2D Ising model at different temperatures.}
Comparison of training-data statistics with equilibrium samples from PCD- and PTT-trained RBMs. \textbf{(a)} Magnetization distributions for $L=20$: PCD exhibits mode collapse at low temperatures, whereas PTT reproduces both modes. \textbf{(b)} Disconnected susceptibility versus $\beta$ for PCD- (down) and PTT-trained (up) models across lattice sizes, compared with the data (black). \textbf{(c)} Number of intermediate models (temperatures) required by PTT (PT) versus $\beta$. \textbf{(d)} Finite-size scaling of the disconnected susceptibility for PTT-trained models using the exact 2D Ising critical exponents. The inset shows the Binder cumulant versus $\beta$, showing the expected crossing at the critical point.
}
    \label{fig:Ising}
\end{figure}
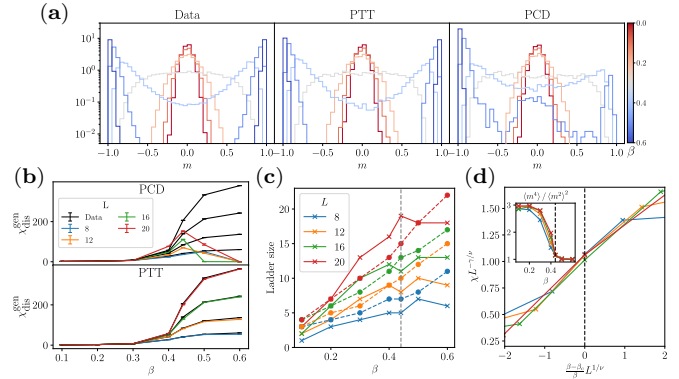
\subsection{Ising 2D}
Several studies have reported that standard RBM training fails to faithfully reproduce the low-temperature distribution of the two-dimensional Ising model~\cite{yevick2021accuracy,gu2022thermodynamics,valle2023capabilities}, despite the existence of an exact RBM representation with only $2L^2$ hidden units~\cite{decelle2024inferring}. As shown below, this limitation arises from inaccurate sampling during conventional CD/PCD training rather than insufficient representational power. To compare PTT and PCD, we train RBMs on the ferromagnetic Ising model on square lattices of size $L\!\times\! L$, with $L\!\in\!\{8,12,16,20\}$, for temperatures spanning both sides of the critical point $\beta_\mathrm{c}\!=\!0.440687$. The distribution evolves from unimodal at high temperatures to bimodal at low temperatures, as shown by the $L=20$ magnetization histograms in Fig.~\ref{fig:Ising}a.

For each $(L,\beta)$, we generate $10^6$ equilibrium configurations using Swendsen--Wang sampling~\cite{swendsen1987nonuniversal} and train RBMs with $N_\mathrm{h}=2L$ hidden units, sufficient for an exact sparse representation~\cite{decelle2024inferring}, using PTT and PCD. At the maximum test log-likelihood, we generate $M_\mathrm{gen}=5000$ equilibrium samples using PTT for PTT-trained models and conventional PT for PCD-trained models, for which a suitable PTT ladder could not be reconstructed \emph{a posteriori}. The PT ladder is built using the same overlap criterion as PTT, and the same thermalization protocol is applied.
We assess generative quality by comparing thermodynamic observables between training and generated samples across $L$ and $\beta$. From the magnetization per spin, $m=L^{-2}\sum_i s_i$, we compute $\langle m^2\rangle$, the disconnected susceptibility $\chi_{\mathrm{dis}}=L^2\langle m^2\rangle$, and the Binder ratio $\langle m^4\rangle/\langle m^2\rangle^2$. Error bars are estimated by bootstrap.

Figure~\ref{fig:Ising}a compares the magnetization distributions of the data and samples generated by PCD- and PTT-trained RBMs. At low temperatures, PCD samples collapse onto a single magnetization sector (negative), whereas PTT samples correctly reproduce both modes. Figure~\ref{fig:Ising}b compares the disconnected susceptibility across lattice sizes: both methods agree in the paramagnetic phase, but PCD deviates sharply near and below the critical temperature, while PTT remains in excellent agreement throughout.
In Fig.~\ref{fig:Ising}d, the disconnected susceptibility of PTT-trained RBMs follows the expected finite-size scaling of the 2D Ising model~\cite{amit2005field}, while the Binder cumulant exhibits the expected crossing at the critical temperature (inset of Fig.~\ref{fig:Ising}d), confirming that PTT reproduces the critical behavior.

Finally, we compare the efficiency of PTT with standard PT, which is expected to perform well for the Ising model, where sampling only requires crossing a second-order transition, unlike more complex multimodal datasets involving first-order barriers~\cite{bereux2025fast}. Figure~\ref{fig:Ising}c shows the number of PTT checkpoints and PT temperature replicas required to reach the same swap acceptance, $\alpha=0.3$, for the same PTT-trained RBMs. Below the critical temperature, both ladder sizes initially increase similarly with $\beta$ and $L$, but at lower temperatures the PTT ladder saturates, while the number of PT replicas continues to grow with $\beta$.

\subsection{Human Genome Dataset} The Human Genome Dataset (HGD)~\cite{yelmen2021creating} is a binary dataset derived from the 1000 Genomes Project~\cite{10002015global}, encoding the presence or absence of mutations relative to a reference genome across 805 genes selected for their high variability between geographical regions. HGD and related variants have been widely used to assess privacy leakage in generative models trained on sensitive genomic data~\cite{yelmen2021creating,yelmen2023deep,yelmen2023overview,szatkownik2025privet}.

\begin{figure}[h!]
    \centering
    \begin{overpic}[width=1\linewidth]{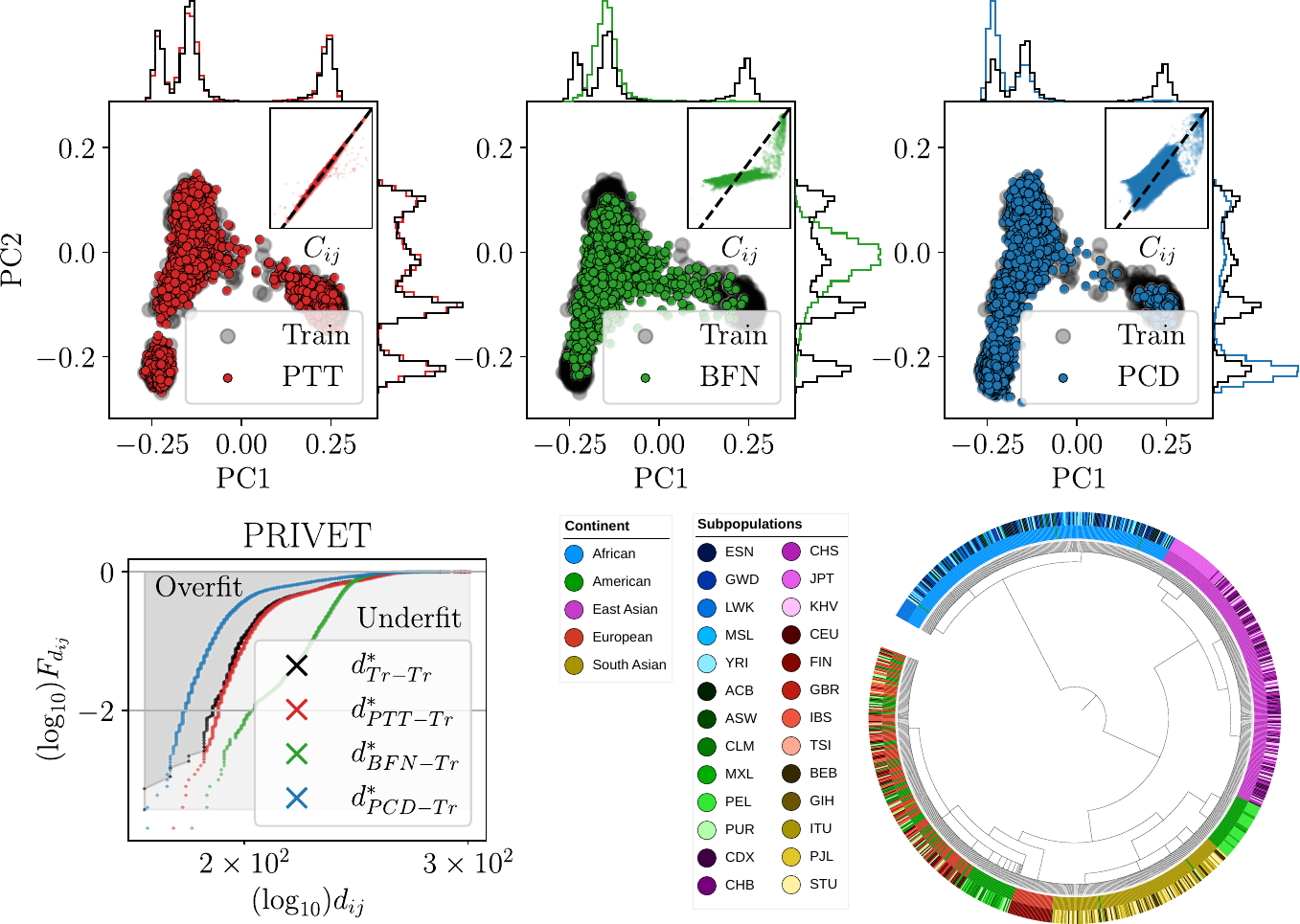}
    \put(2,65){{\figpanel{a}}}
    \put(3,30){{\figpanel{b}}}
    \put(37,30){{\figpanel{c}}}
    \end{overpic}
    \caption{\textbf{Comparison The same behavior is recovered with AIS usingof the sampling quality on HGD.} \textbf{(a)} Scatter plots of the training dataset and the samples generated by different models projected onto the first two principal components of the dataset. The inset shows scatter plots of the empirical two-body correlations measured on the training dataset (x-axis) and on the generated samples (y-axis). The black dashed line indicates the identity. \textbf{(b)} Distribution of nearest-neighbor distances between generated samples and the training dataset. \textbf{(c)} Hierarchical clustering of test samples based on their closest free-energy minima across training of the PTT model, following Ref.~\cite{decelle2023unsupervised}. Leaves denote test samples and are colored by continental origin (inner ring) and subpopulation (outer ring), revealing clear organization across continents and several subpopulations.
    }
    \label{fig:HGD}
\end{figure}

This dataset exhibits strong clustering along its first principal components, Fig.~\ref{fig:HGD}a, making training with conventional sampling particularly difficult~\cite{bereux2023learning,szatkownik2024diffusion,bereux2025fast}; its limited sample size further makes it a challenging benchmark for generative models, as shown in previous studies~\cite{yelmen2021creating,yelmen2023deep}.
Early in maximum-likelihood training, the phase space fragments into well-separated clusters, trapping Gibbs sampling and causing PCD to misestimate their relative weights. PTT instead exploits the continuity of the model distribution along the training trajectory to sample intermediate models efficiently. 

We first compare samples generated by PTT- and PCD-trained RBMs with those from BFN, a state-of-the-art method for discrete data generation, in Fig.~\ref{fig:HGD}a. In each case, we select the model with the highest test log-likelihood and project generated samples onto the first two principal components of the training data (black points), with marginal one-dimensional histograms shown alongside. Both BFN and PCD either exhibit mode collapse or fail to reproduce all modes of the data distribution, and neither accurately captures the pairwise statistics (insets of Fig.~\ref{fig:HGD}a). By contrast, the PTT-trained RBM reproduces both the multimodal structure and pairwise statistics with high accuracy.

We next use the PRIVET method~\cite{szatkownik2025privet} to quantify underfitting, overfitting, and memorization. PRIVET compares the distribution of nearest-neighbor (NN) distances between generated samples and the training set with the distribution of NN distances within the training set itself, which serves as the reference distribution~\cite{szatkownik2025privet} (More details are given in the Appendix). NN distances systematically smaller than the reference indicate overfitting and memorization, whereas larger distances indicate underfitting. The objective is therefore to reproduce the reference distribution.
As shown in Fig.~\ref{fig:HGD}b, the BFN underfits the data, while the PCD-trained RBM, owing to its mode collapse, strongly overfits the corresponding cluster. By contrast, the PTT-trained RBM almost perfectly matches the reference distribution at all distances.

To further investigate the structure learned by the RBM, we analyze the organization of the free-energy landscape using the hierarchical clustering procedure introduced in Ref.~\cite{decelle2023unsupervised}. As shown in Fig.~\ref{fig:HGD}c, samples assigned to the same free-energy minimum cluster together, revealing a clear hierarchical organization that first separates continental populations and subsequently resolves finer subpopulation structure. This indicates that the learned energy landscape captures meaningful genetic relationships without using any label information.

\subsection{Protein Family Sequence Data}
\onecolumngrid
\begin{figure*}[t]
    \centering
    \begin{overpic}[width=0.7\linewidth]{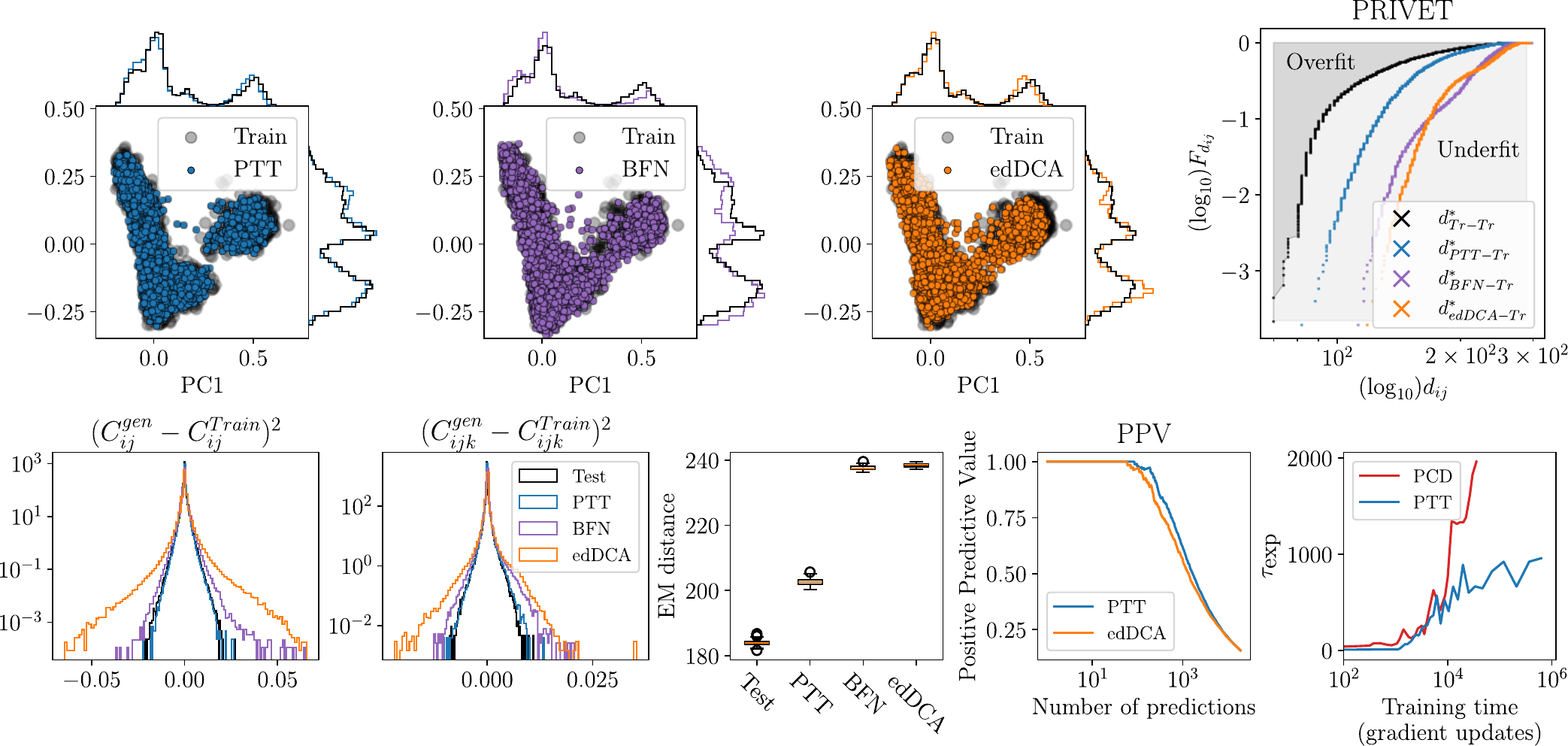}
    \put(3,45){{\figpanel{a}}}
    \put(0,20){{\figpanel{b}}}
    \put(43,20){{\figpanel{c}}} 
    \put(76,48){{\figpanel{d}}} 
    \put(63,20){{\figpanel{e}}} 
    \put(82,20){{\figpanel{f}}} 
    \end{overpic}
    \caption{\textbf{$\beta$-Lactamase (PF13354).} \textbf{(a)} Generated and training samples projected onto the first two principal components. \textbf{(b)} Distribution of errors in two- and three-body correlations relative to the training data. \textbf{(c)} Earth Mover Distance between generated and training samples. \textbf{(d)} Nearest-neighbor distance distributions to the training set. \textbf{(e)} Positive Predictive Value (PPV) as a function of the number of top-ranked pairwise couplings. \textbf{(f)} Evolution of $\tau_{\exp}$ as a function of training time for PCD- and PTT-trained models.}
    \label{fig:PF13354}
\end{figure*}
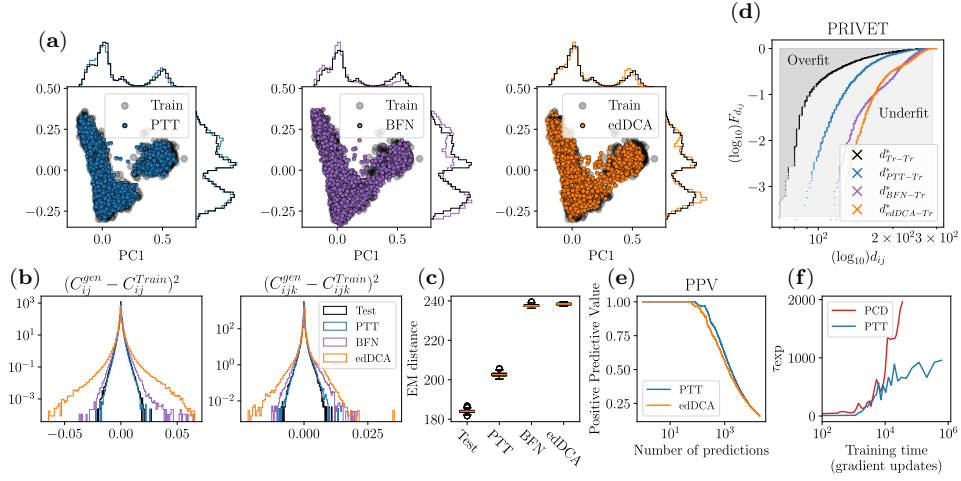
\twocolumngrid
We next consider a multiple sequence alignment (MSA) of homologous sequences from the beta-lactamase domain family \href{https://www.ebi.ac.uk/interpro/entry/pfam/PF13354/}{PFAM\_ID:PF13354}, whose strongly clustered distribution makes EBM training particularly challenging, with potentially very long thermalization times~\cite{muntoni2021adabmdca}. Pairwise EBMs, often referred to as direct coupling analysis (DCA) models, have long been used to generate protein sequences and to infer residue--residue couplings that are predictive of contacts in three-dimensional structures.
RBMs generalize this framework by capturing effective multibody interactions~\cite{decelle2025inferring}. This dataset therefore provides a stringent benchmark of both generative accuracy and interaction inference. We show that PTT-trained RBMs better reproduce the empirical sequence distribution than competing models while improving contact prediction over edDCA~\cite{barrat2021sparse,rosset2026adabmdca}.

We compare samples generated by PTT-trained RBMs, BFN, and edDCA after projection onto the first two principal components of the data in Fig.~\ref{fig:PF13354}a. All three models recover the main clusters, although BFN and edDCA reproduce their relative weights less accurately. To quantify these differences, we examine the distributions of the errors in two- and three-body correlations, Fig.~\ref{fig:PF13354}b. Despite its qualitatively accurate low-dimensional projection, edDCA shows substantially larger correlation errors than BFN, while the PTT-trained RBM outperforms both and approaches the test-set baseline. Consistently, the Earth Mover Distance~\cite{rubner1998metric} between generated and training samples, Fig.~\ref{fig:PF13354}c, is lowest for PTT, although it remains above the test--train reference value. The same trend appears in the nearest-neighbor distance distribution Fig.~\ref{fig:PF13354}d: all models remain in the underfitting regime, but PTT lies significantly closer to the data reference.

Finally, we extract the effective two-body interactions learned by the RBM from its weights~\cite{decelle2025inferring} and compare the resulting couplings with the PF13354 contact map. We restrict the comparison to edDCA, for which pairwise couplings are explicit model parameters. Performance is quantified by the Positive Predictive Value (PPV), i.e., the fraction of true contacts among the strongest predicted couplings. Both methods recover true contacts at the top of the ranking, but as more couplings are included, the PPV of edDCA decreases more rapidly, while the PTT-trained RBM retains higher predictive accuracy.

\subsection{Neural recordings} 
We next consider Neuropixels recordings from an Allen Institute experiment~\cite{doi:10.1126/science.abf4588}, comprising the simultaneous activity of 2,028 neurons across cortical and subcortical regions in mice performing a visual change-detection task with familiar and novel images. Spike trains are binned into 20 ms intervals and binarized according to whether each neuron fired at least once; preprocessing details are given in Ref.~\cite{bereux2026uncovering}. A key challenge is the strong temporal structure of this dataset, which induces correlations between consecutive samples, substantially reducing the effective number of independent observations, while generating a highly multimodal distribution that makes training particularly difficult.

\begin{figure}
    \centering
    \begin{overpic}[width=\linewidth]{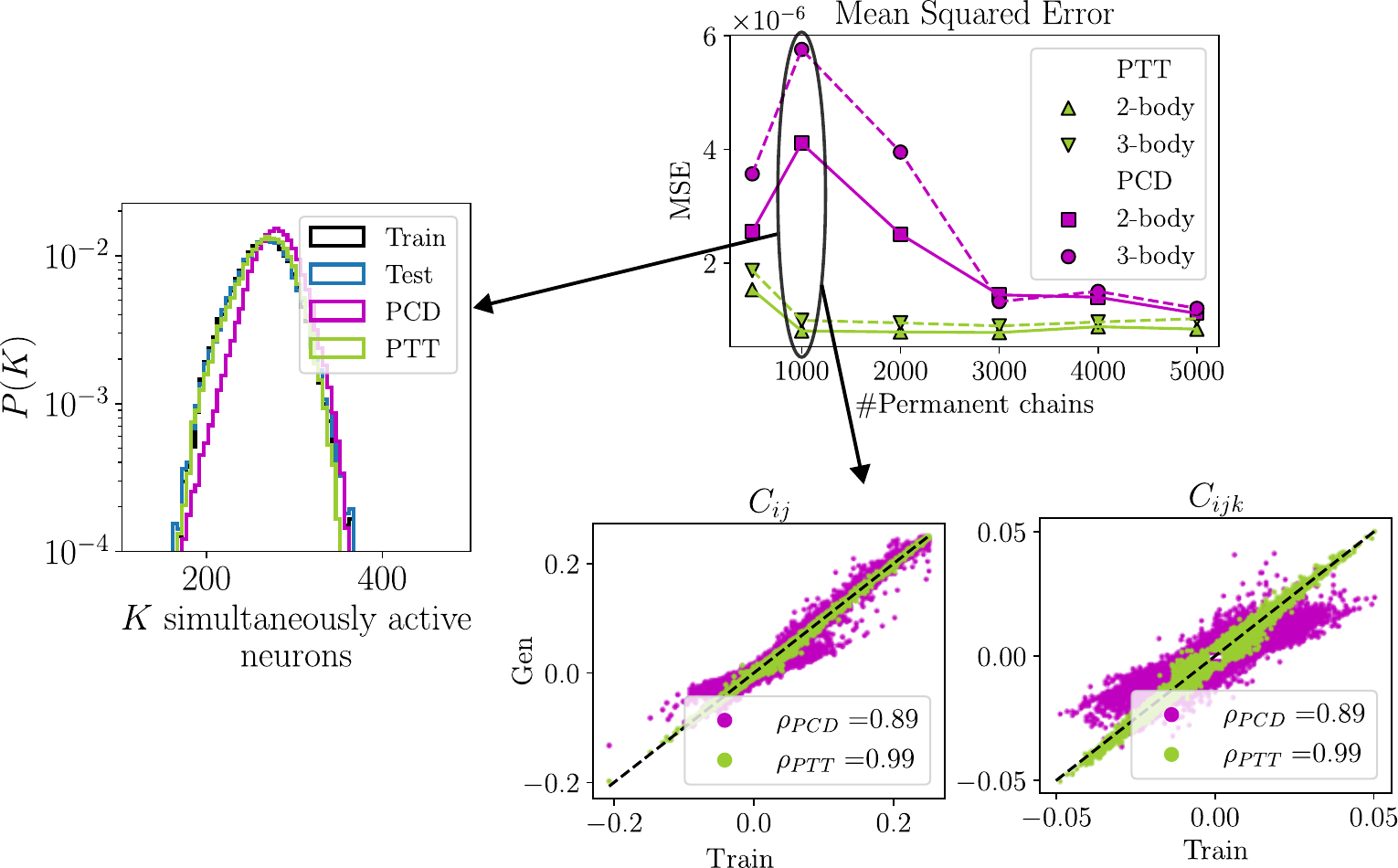}
    \put(48,63){{\figpanel{a}}}
    \put(3,50){{\figpanel{b}}}
    \put(38,27){{\figpanel{c}}}
    \end{overpic}
    \caption{\textbf{Neural activity}. Impact of the number of persistent chains during training.
\textbf{(a)} MSE of two- and three-body correlations for PCD- and PTT-trained RBMs versus the number of parallel chains used to estimate the likelihood gradient during training.
\textbf{(b)} Distribution of the number of simultaneously active neurons, $K=\sum_i n_i$, for models trained with $1000$ chains.
\textbf{(c)} Two- and three-body correlations of generated versus training samples for models trained with $1000$ chains. The dashed line denotes identity; Pearson correlations are reported in the legend. All statistics are computed by comparing $15{,}000$ samples from each dataset.} 
    \label{fig:comp_n_chains}
\end{figure}

We find that PCD performance depends strongly on training hyperparameters, in particular the number of parallel chains and the minibatch size used to estimate the likelihood gradient. To illustrate this, we train identical RBMs with PCD and PTT using $1000$ parallel chains, a standard choice in practice and the value used throughout this work. As shown in Fig.~\ref{fig:comp_n_chains}a, PCD fails to reproduce the two- and three-point correlations of the data, whereas PTT accurately captures them. Figure~\ref{fig:comp_n_chains}b compares the distribution $P(K)$ of the number $K$ of simultaneously active neurons, a nonlinear observable sensitive to higher-order statistics. PCD exhibits a clear shift in the high-probability region relative to the train/test data, while PTT closely matches the empirical distribution, with only a small deviation in the low-probability tail.

In Fig.~\ref{fig:comp_n_chains}, we train identical RBMs with PCD and PTT while varying only the number of parallel chains used to estimate the negative phase. Performance is quantified by the mean-squared error between the two- and three-body correlations of generated and training samples. PTT remains stable across all chain counts and consistently achieves lower errors. By contrast, PCD degrades markedly at small numbers of chains and approaches PTT performance only when using about $5000$ chains.

\subsection{Medical Recommendation Data}

As a final benchmark, we consider the Medical Recommendation Dataset~\cite{medical_data}, a tabular dataset containing 241 patient profiles described by symptoms, causes, diagnosed illnesses, and recommended treatments.The dataset is moderately sparse and combines heterogeneous categorical variables with structured dependencies between symptoms, diagnoses, and treatments.  It therefore provides a representative benchmark for tabular generative modeling. Modeling tabular data remains challenging for modern generative models due to heterogeneous features, imbalanced and multimodal marginals, and complex cross-variable dependencies~\cite{xu2019modeling,kotelnikov2023tabddpm}. 

\begin{figure}
    \centering
   \begin{overpic}[width=\linewidth]{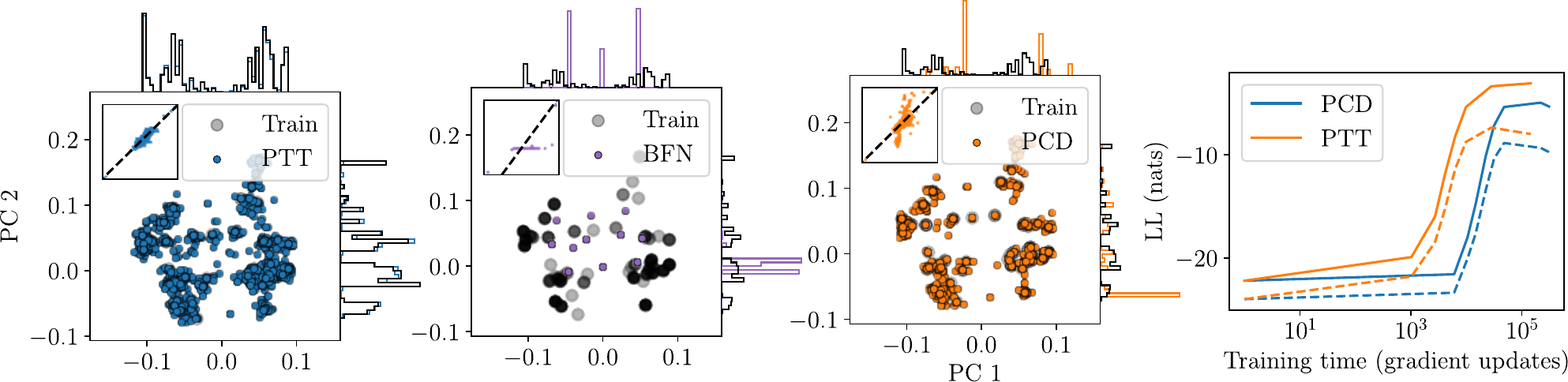}
    \put(3,20){{\figpanel{a}}}
    \put(73,20){{\figpanel{b}}}
    \end{overpic}
    \caption{\textbf{Sampling quality} on Medical data. \textbf{(a)} Scatter plots of generated data against training data on the first two principal components. The inset shows the two-body correlations estimated on the generated data against the train dataset. \textbf{(b)} Train (full line) and test (dashed line) log-likelihood of the PCD- and PTT-trained model during training. }
    \label{fig:medical}
\end{figure}

We train RBMs with PCD and PTT, together with a BFN model, on this dataset. In order to binarize the dataset, each of the categorical variable was one hot encoded in order to binarize the dataset. As shown in Fig.~\ref{fig:medical}, PTT reaches a substantially higher test log-likelihood than PCD while avoiding the undesirable dynamical behavior observed during PCD training. We also project generated samples onto the first two principal components of the data. While PTT reproduces the main structure of the empirical distribution, BFN fails to generate realistic samples and does not capture the observed modes.

\section{Conclusion}

In this work, we introduced a training algorithm based on an optimized sampling scheme inspired by parallel tempering and designed to exploit the learning trajectories of EBMs, which have recently been shown to undergo second-order phase transitions during training. The method enables high-precision equilibrium training and allows compact RBMs to outperform more advanced generative models across a broad range of scientific datasets, despite their substantially smaller architectures and lower training cost.

Although EBM training has long been regarded as a major challenge, we showed that our approach remains reliable even for strongly multimodal and clustered distributions and in low-data regimes. Beyond stable optimization, it enables accurate log-likelihood estimation, allowing controlled studies of convergence and hyperparameter selection, including the number of hidden units and early stopping. It also provides direct diagnostics of sampling quality, since loss of equilibration can be detected from the replica-exchange dynamics.

We expect these advances to facilitate the broader use of EBMs in scientific and applied settings, where their compactness, interpretability, and access to accurate likelihood estimates provide distinctive advantages over many alternative generative approaches.

\section*{Data Availability}
The code used to train RBMs with PTT, together with notebooks and all the data required to analyze the trained models and reproduce the figures in this paper, will be made available at \url{https://www.github.com/DsysDML/ptt_paper}.

\begin{acknowledgments} The authors thank Stefano Bae and Lorenzo Rosset for useful discussions regarding the code and its applications, and acknowledge financial support from grant PID2024-158623NB-C21, funded by MICIU/AEI/10.13039/501100011033 and the ERDF/EU.
This work is supported by the European Research Council (ERC) under the European Union's Horizon Europe research and innovation programme (Grant Agreement No. 101231393 - BeME).
\end{acknowledgments}

\bibliography{ref}

\appendix
\section*{Appendix}

\subsection*{Hyperparameters} All models are trained using two PTT sweep attempts between the reservoir and the last two checkpoints in the ladder. Each PTT sweep consists of attempting swaps of configurations between adjacent models, followed by 10 alternating Gibbs updates, during which the visible and hidden layers are updated successively. The learning rate is initialized at $10^{-3}$ and is automatically adjusted throughout training. New checkpoints are added to the ladder whenever the swap acceptance rate between the reservoir and the last checkpoint falls below $\alpha = 0.3$. Each time a new checkpoint is added, a PTT simulation involving the entire ladder is run until thermalization is reached (see below for the thermalization criterion), after which a new equilibrium reservoir is generated. The swap acceptance rates across the full PTT ladder are continuously monitored. If any of them falls below $\alpha - 0.2$, training is stopped, as this indicates that the Markov chains have likely drifted out of equilibrium. In this case, the maximum learning rate should be reduced before restarting training.

\subsection*{Adaptive learning rate}
The parameters of the EBM are optimized by stochastic gradient ascent. Since PTT provides low-noise gradient estimates once the sampler has equilibrated, the relative orientation of successive gradients carries useful information about the optimization dynamics. We exploit this information to adapt the learning rate during training.

Let \(\mathbf{g}_t\) denote the gradient of the log-likelihood at optimization step \(t\). We quantify the alignment between two consecutive gradients through their cosine similarity,
\begin{equation}
\mathrm{cossim}(\mathbf{g}_t,\mathbf{g}_{t-1})
=
\frac{\mathbf{g}_t^\top\mathbf{g}_{t-1}}
{\|\mathbf{g}_t\|\,\|\mathbf{g}_{t-1}\|}.
\end{equation}
When consecutive gradients are well aligned (\(\mathrm{cossim}\approx1\)), the optimization consistently follows the same direction, indicating that a larger learning rate can safely accelerate convergence. 
Conversely, negative cosine similarity signals that successive updates cancel each other out, indicating instability and the need to reduce the learning rate. A value close to zero suggests orthogonal updates, meaning there is no clear direction to follow, and the learning rate should remain stable.

We therefore adjust the learning rate dynamically according to the measured cosine similarity, increasing it when gradients remain aligned, decreasing it when they  become anti-aligned and keeping it stable when the updates are decorrelated. Finally, in the case of RBMs, the cosine similarity is computed separately for the visible bias, hidden bias and weight matrix to avoid a noisy gradient in one component disrupting the others.

An additional safeguard is provided by the PTT dynamics itself. The swap acceptance between neighboring checkpoints measures the overlap of their equilibrium distributions. If it falls below a prescribed threshold, that we fix always to $\alpha=0.3$, the optimization step is rejected, the parameters are restored to the previous checkpoint, and the learning rate is halved before training resumes. This prevents overly large parameter updates from disrupting the PTT ladder while automatically adapting the learning rate to the local optimization landscape.

\subsection*{Log-likelihood estimation}
Beyond substantially accelerating sampling, PTT provides, at essentially no additional computational cost, an accurate estimate of the log-likelihood throughout and after training~\cite{bereux2025fast}, provided that the replica ensemble has equilibrated. The key observation is that the partition functions of two consecutive models in the ladder satisfy
\begin{align}
Z_t
&=
\sum_{\bm{v}}
e^{-E_t(\bm{v})}
=
\sum_{\bm{v}}
e^{-E_{t-1}(\bm{v})}
e^{E_{t-1}(\bm{v})-E_t(\bm{v})}
\nonumber\\
&=
Z_{t-1}
\left\langle
e^{E_{t-1}-E_t}
\right\rangle_{t-1},
\end{align}
where \(\langle\cdot\rangle_{t-1}\) denotes an expectation over the equilibrium distribution of model \(t-1\). Starting from the first checkpoint, whose partition function is known analytically, the partition function of every subsequent model is obtained recursively as
\begin{equation}
\log Z_t
=
\log Z_{t-1}
+
\log
\left\langle
e^{E_{t-1}-E_t}
\right\rangle_{t-1}.
\end{equation}
The expectation is estimated using the equilibrium configurations already generated by PTT. Since neighboring checkpoints are introduced only when their overlap is sufficiently large, the energy difference \(E_t-E_{t-1}\) remains small and the exponential reweighting factor has low variance. 

Once \(\log Z_t\) is known, the average log-likelihood of a dataset \(\mathcal D=\{\bm{v}^\mu\}_{\mu=1}^{M}\) follows directly from
\begin{equation}
\mathcal{L}_t
=
\frac{1}{M}
\sum_{\mu=1}^{M}
\log p_t(\bm{v}^\mu)
=
-\frac{1}{M}
\sum_{\mu=1}^{M}
E_t(\bm{v}^\mu)
-
\log Z_t.
\end{equation}

Furthermore, PTT also provides a simple diagnostic of thermalization by monitoring the diffusion of replicas across the ladder~\cite{alvarez2010nature} (see Appendix).

\subsection*{Thermalization criterion and reservoir creation}

To ensure equilibrium sampling, we follow the approach of Ref.~\cite{banos2010nature}. We run \(n_{\mathrm{chains}}=1000\) independent PTT simulations (in this work referred to as \emph{parallel chains}), each containing \(N_{\mathrm{m}}\) replicas evolving simultaneously at different model checkpoints. During the simulation, replicas diffuse along the ladder through swap moves. In equilibrium, every checkpoint is visited with equal probability. Denoting by \(n_i(t)\) the checkpoint occupied by replica \(i\) at Monte Carlo time \(t\), we compute the normalized autocorrelation function
\begin{equation}
C(t)=
\frac{\left\langle \left(n_i(t+t_0)-\langle n\rangle\right)
\left(n_i(t_0)-\langle n\rangle\right)\right\rangle}
{\left\langle\left(n_i(t_0)-\langle n\rangle\right)^2\right\rangle},
\end{equation}
where \(\langle\cdot\rangle\) denotes an average over all replicas, parallel chains, and time origins \(t_0\), and
\(\langle n\rangle=(N_\mathrm{m}-1)/2\). At long times,
\begin{equation}
C(t)\sim A\,e^{-t/\tau_{\mathrm{exp}}},
\end{equation}
which defines the \textit{exponential autocorrelation time} \(\tau_{\mathrm{exp}}\), corresponding to the slowest relaxation mode of the random walk through the ladder. Whenever a new checkpoint is added, the \(n_{\mathrm{chains}}\) parallel chains are first evolved for at least \(20\,\tau_{\mathrm{exp}}\) to ensure thermalization.

After thermalization, statistically independent configurations are generated by recording one configuration every \(2\tau_{\mathrm{int}}\) Monte Carlo steps, where the integrated autocorrelation time is obtained from the self-consistent relation
\begin{equation}
\tau_{\mathrm{int}}
=
\frac{1}{2}
+
\sum_{t=0}^{6\tau_{\mathrm{int}}}
C(t).
\end{equation}
Whenever a new checkpoint is added to the PTT ladder during training, this protocol is used to construct a reservoir of \(N_{\mathrm{res}}=10N_\mathrm{chains}\) thermalized and statistically independent configurations. The same thermalization and sampling protocol is used throughout this work to generate all reported samples.

\subsection*{PRIVET assessment of under- and overfitting}
To assess under- and overfitting, we use PRIVET~\cite{szatkownik2025privet}, a nearest-neighbor method based on extreme-value statistics. For a sample $x$ and a reference set $\mathcal{D}$, we define the nearest-neighbor distance
\[
\delta(x,\mathcal{D})=\min_{y\in\mathcal{D}} d(x,y).
\]
We compare the distribution of $\delta(x,\mathcal{D}_{\rm train})$ for generated samples with a reference distribution obtained from distances between two disjoint subsets of the real data. If generated samples are systematically farther from the training data than this reference, the model underfits; if both distributions agree, the model is consistent with good generalization; if generated samples are systematically closer, this indicates overfitting and possible memorization. PRIVET formalizes these deviations using an extreme-value fit to the nearest-neighbor distribution, allowing statistically significant excesses of unusually small distances to be identified.

\subsection*{Problems with PCD in data-scarce regimes}

Here we illustrate the failure of PCD-trained EBMs in data-scarce regimes using binarized MNIST~\cite{lecun1998gradient}. While PCD performs well on the full $50{,}000$-sample training set, reducing the dataset size exposes severe sampling pathologies.

\begin{figure}[h!]
\begin{overpic}[width=1\linewidth,trim=0 250 0 150,clip]{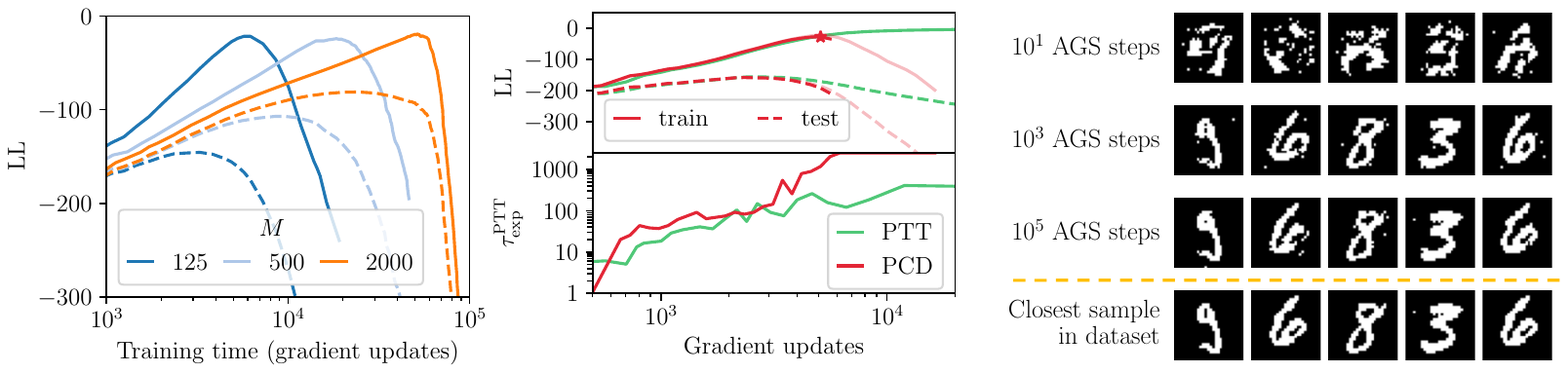}
    \put(0,25){{\figpanel{a}}}
    \put(39,25){{\figpanel{b}}}
    \put(39,12){{\figpanel{c}}}
    \put(62,25){{\figpanel{d}}}
    \end{overpic}
\caption{\textbf{Failure modes of PCD training in data-scarce regimes.}
(a) Training (solid) and test (dashed) log-likelihoods during PCD training for different dataset sizes $M$.
(b) For $M=100$, comparison between PCD (red, learning rate $0.01$) and PTT (green, adaptive learning rate with $0.01$ maximum learning rate). Red stars mark the onset of the dynamical slowdown, where thermalization requires more than $5\times10^4$ PTT sweeps, according to our $20\tau_{\mathrm{exp}}$ criterion. Dark curves use thermalized PTT estimates, whereas shaded curves use short 1000-sweep runs.
(c) Corresponding exponential autocorrelation times $\tau_{\mathrm{exp}}$ along training.
(d) Five independent alternating Gibbs trajectories of $10^5$ sweeps, initialized from the random configurations shown on the left and using the PCD model marked by the red star in (b). The trajectories rapidly become trapped near individual training examples and remain confined there.
}
    \label{fig:MNIST}
\end{figure}

We train PCD-RBMs on subsets of size $M$ and save checkpoints throughout training. These checkpoints are assembled a posteriori into a PTT ladder like in Ref.~\cite{bereux2025fast} to accelerate sampling, assess thermalization, and estimate the log-likelihood. Figure~\ref{fig:MNIST}a shows the training and test log-likelihoods estimated from short 1000-sweep PTT runs. Both eventually decrease sharply, including the training likelihood despite it being the optimized objective. The same behavior is recovered with AIS using sufficiently many chains and intermediate temperatures. We also verified this behavior for RBMs with sufficiently few hidden units to allow exact log-likelihood computation by enumeration, confirming that the train-LL collapse persists in low-data regimes.
By contrast, PTT-trained RBMs display standard overfitting, with the training likelihood continuing to increase after the test likelihood decreases (Fig.~\ref{fig:MNIST}b).

The breakdown is accompanied by a sharp growth of the exponential autocorrelation time $\tau_{\mathrm{exp}}$ (Fig.~\ref{fig:MNIST}c). Beyond the red star, equilibration cannot be achieved within $5\times10^4$ PTT sweeps, whereas relaxation remains controlled for PTT-trained models. Figure~\ref{fig:MNIST}d further shows that the freezing corresponds to trajectories becoming trapped near individual training examples and generating memorized copies after only a few Gibbs steps. This sudden arrest in the relaxation dynamics is commonly observed in PCD-trained models as we show for instance in Fig.~\ref{fig:PF13354}--f for proteins.

\end{document}

%% file: math_commands.tex
\usepackage{amsmath,amsfonts,bm}
\usepackage{xcolor}

\def\eqref#1{equation~\ref{#1}}

\def\1{\bm{1}}

\DeclareMathAlphabet{\mathsfit}{\encodingdefault}{\sfdefault}{m}{sl}
\SetMathAlphabet{\mathsfit}{bold}{\encodingdefault}{\sfdefault}{bx}{n}

\newcommand{\mc}[1]{\mathcal#1}

\newcommand{\mean}[1]{\left\langle #1\right\rangle}